\documentclass[runningheads]{llncs}
\usepackage{wrapfig}
\usepackage{graphicx}
\usepackage{float}
\PassOptionsToPackage{hyphens}{url}\usepackage{hyperref}
\usepackage{color,soul}
\usepackage{tikz}
\usetikzlibrary{calc,trees,positioning,arrows,chains,shapes.geometric,%
    decorations.pathreplacing,decorations.pathmorphing,shapes,%
    matrix,shapes.symbols}

\tikzset{
>=stealth',
  punktchain/.style={
    rectangle, 
    rounded corners, 
    draw=black, very thick,
    text width=20em, 
    minimum height=1em, 
    text centered, 
    on chain},
  line/.style={draw, thick, <-},
  element/.style={
    tape,
    top color=white,
    bottom color=blue!50!black!60!,
    minimum width=8em,
    draw=blue!40!black!90, very thick,
    text width=10em, 
    minimum height=3.5em, 
    text centered, 
    on chain},
  every join/.style={->, thick,shorten >=1pt},
  decoration={brace},
  tuborg/.style={decorate},
  tubnode/.style={midway, right=2pt},
}

\tikzset{
  treenode/.style = {shape=rectangle, rounded corners,
                     draw, align=center,
                     top color=white, bottom color=white},
  root/.style     = {treenode, font=\Large, bottom color=white},
  env/.style      = {treenode, font=\normalsize},
  dummy/.style    = {circle,draw}
}

\begin{document}

\title{
  Electric Sheep Team Description Paper Humanoid League Kid-Size 2019
  \thanks{
    Supported by the University of Canterbury \href{https://www.canterbury.ac.nz/engineering/}{College of Engineering }\&
    \href{https://www.hitlabnz.org/}{HIT Lab NZ}
  }
}

\titlerunning{Electric Sheep Team Description Paper 2019}

\author{
  Dan Barry\inst{1} \and
  Andrew Curtis-Black\inst{1} \and
  Merel Keijsers\inst{1} \and
  Munir Shah\inst{2} \and
  Matthew Young\inst{1} \and
  Humayun Khan\inst{1} \and
  Banon Hopman\inst{1}
}

\authorrunning{D. Barry et al.}

\institute{
  University of Canterbury, Christchurch, New Zealand \and
  Stats NZ  \\
  \url{https://humanoid.science}
}

\maketitle
\begin{abstract}
In this paper we introduce the newly formed New Zealand based RoboCup Humanoid Kid-Size team, Electric Sheep. We describe our developed humanoid robot platform, particularly our unique take on the chassis, electronics and use of several motor types to create a low-cost entry platform. To support this hardware, we discuss our software framework, vision processing, walking and game-play strategy methodology. Lastly we give an overview of future research interests within the team and intentions of future contributions for the league and the goal of RoboCup.
\end{abstract}


\section{Introducing Electric Sheep}
\label{sec:introduction}

\begin{wrapfigure}{i}{0.20\textwidth}
  \vspace{-2.5em}
    \includegraphics[width=0.17\textwidth]{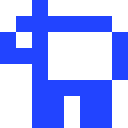}
\end{wrapfigure}
The team Electric Sheep was founded mid-2017 in Christchurch as the first New Zealand humanoid RoboCup team. The name is derived both from the novel by Phillip K. Dick \emph{Do Androids Dream of Electric Sheep?} and the University of Canterbury's historic background in agriculture \cite{UC2018}.

Funding was acquired from the College of Engineering and the HITlab NZ in early 2018, which allowed the team to start building their low-cost humanoid platform with the intention of open-sourcing the designs after the competition. Although the team is new, it also has previous RoboCup experience from the captain who is from the UK-based team \emph{Bold Hearts}, which notably achieved 2nd place in the world championship kid-sized competition in 2014 \cite{ScheunemannDijkRossiBarryPolani2018}.

Electric Sheep will compete in the kid-sized humanoid league, which the team believe to have the lowest cost boundary to entry, a league offering a large number of interesting competitors and opportunities for future collaboration and improvement. Our intention is to combine the multiple backgrounds of the team members and foster interest in robotics in New Zealand.

\section{Architecture}
\label{sec:architecture}


\subsection{Chassis}
\label{subsec:chassis}

\begin{figure}[!htb]
  \centering
  \minipage{0.32\textwidth}
    \includegraphics[width=\linewidth]{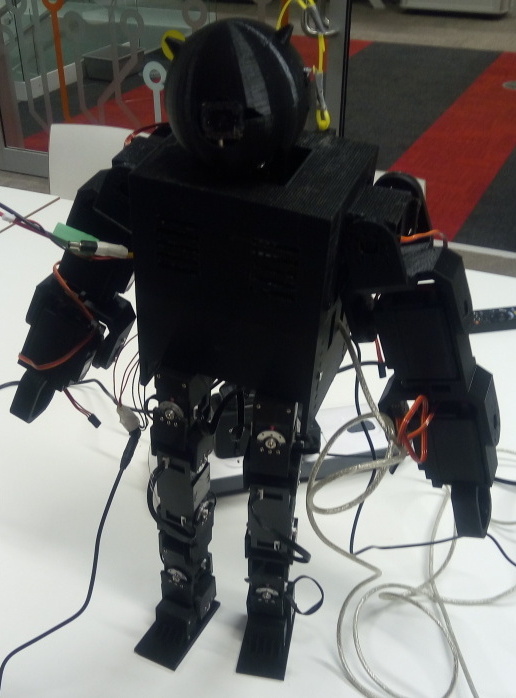}
    \caption{Humanoid platform.}
    \label{fig:robot}
  \endminipage
  \hfill
  \minipage{0.32\textwidth}
    \includegraphics[width=\linewidth]{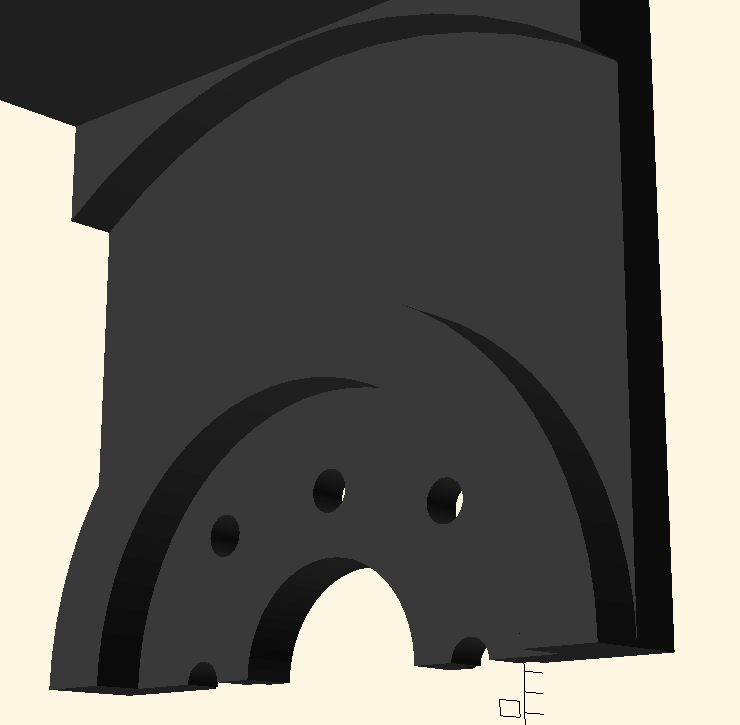}
    \caption{SM30 horn mount.}
    \label{fig:horn}
  \endminipage
  \hfill
  \minipage{0.32\textwidth}
    \includegraphics[width=\linewidth]{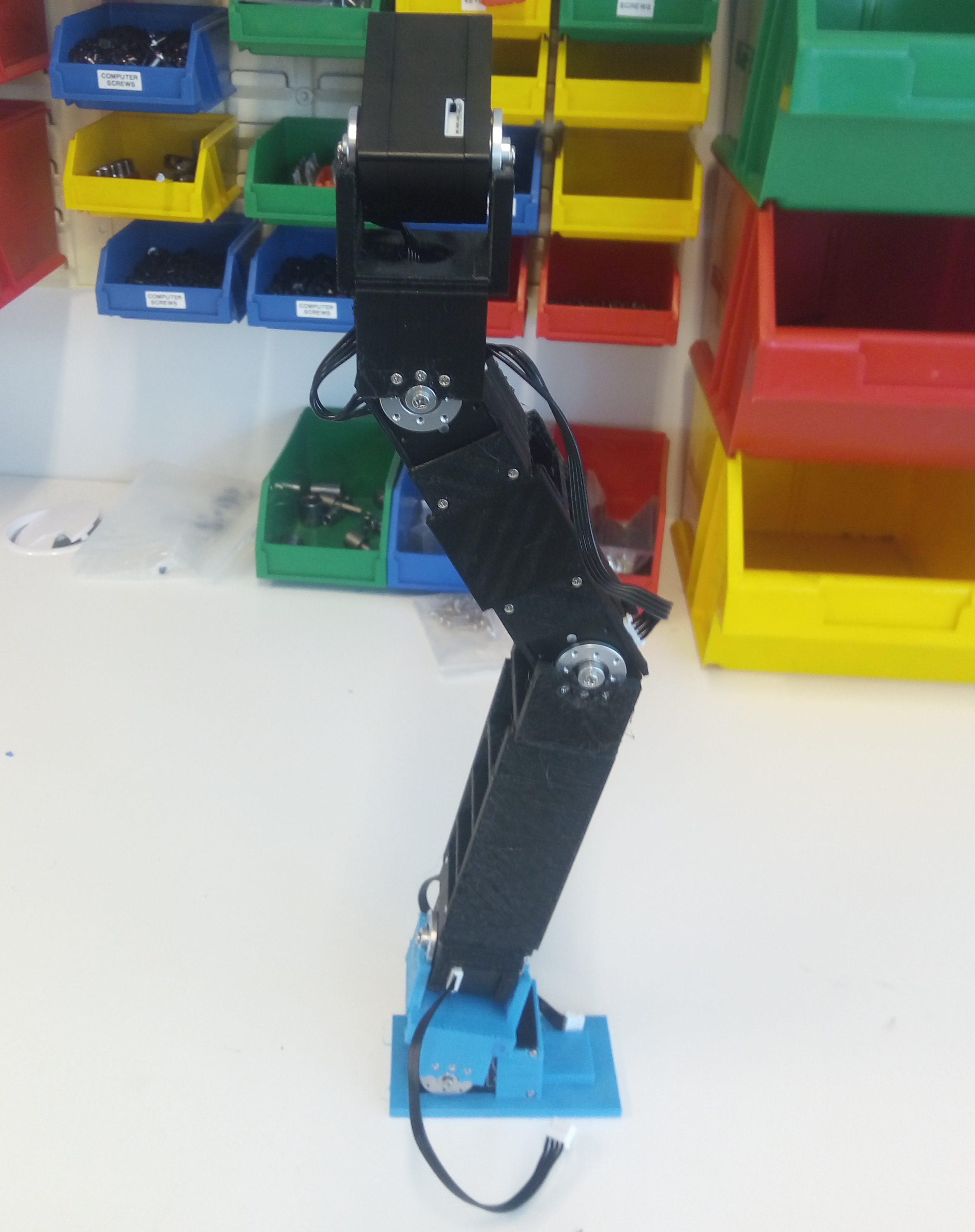}
    \caption{3D printed leg.}
    \label{fig:leg}
  \endminipage
\end{figure}

The robotic platform was designed and built by the team over a period of 8 months. Part of our time has been spent developing a unique humanoid platform in the parametric CAD design software: OpenSCAD (see Figure~\ref{fig:horn}) \cite{Kintel2018} and  3D printing with a MakerBot Replicator 2. Figure~\ref{fig:robot} shows the robots' chassis which consists of 22 3D printed parts; 6 for each leg, 3 for each arm, and two for both the body and the head (not including internal PCB mounts). Our platform thus is easily configurable, printable and simulatable.

\subsection{Electronics}
\label{subsec:electronics}

A Raspberry Pi 3 B+ running Raspbian lite is used as the main board and is responsible for processing all inputs and outputs of the platform. This was chosen as a powerful, low-cost, well documented and understood computing board.

Our platform is powered by a 4 cell 14.8V 2200mAh LiPo battery which is reduced down to 12V (smart motors) and also converted to 5V (main board, control boards, PWM servos). A 4 cell LiPo was chosen over a 3 cell LiPo as the voltage remains above 12V throughout operation, which allows the smart servos to operate with a reliable torque characteristic, where decrease in voltage would result in decrease in torque during operation. 

\subsection{Motors}
\label{subsec:motors}

\begin{figure}
  \centering
  \includegraphics[height=5cm]{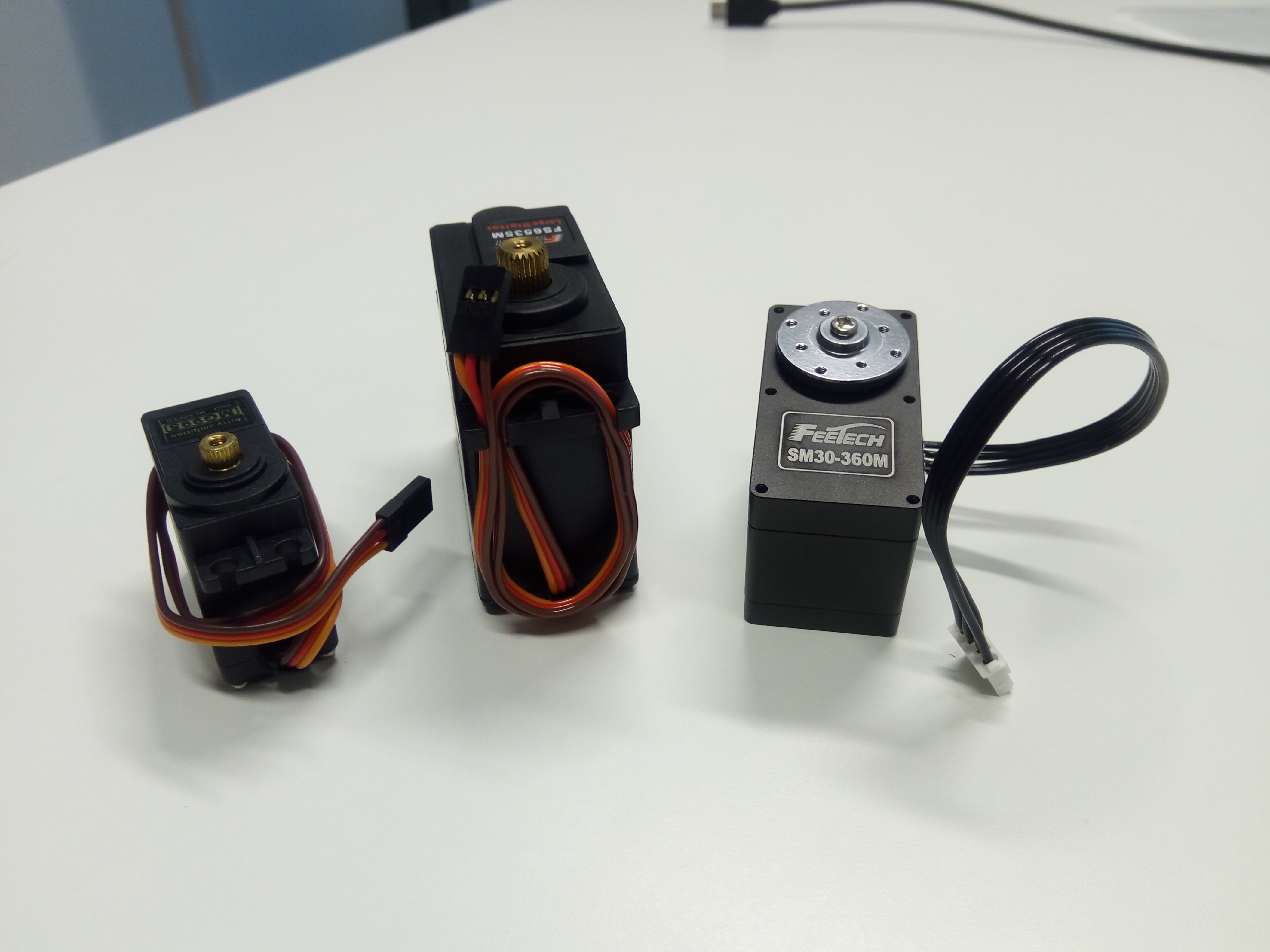}
  \caption{Left: MG995, Center: FeeTech FS6535, Right: FeeTech SM30.}
  \label{fig:motors}
\end{figure}

Three different motors are used for the head, arm, and leg movements (as shown left in Figure~\ref{fig:motors}). To integrate the signals from these motors, two different controller boards are used to control motors via serial and PWM, controlled by the main board via serial and I\textsuperscript{2}C. The smart motors are placed in the legs, as shown in Figure~\ref{fig:leg}.

The robots have two degrees of freedom (DoF) in their ankles, one DoF in their knees, three DoF in the hip joints, two DoF in the shoulder joints, two DoF in the elbow joints, and two DoF in the neck; for a total of 20 joints.

\section{Software}
\label{sec:software}


\subsection{Framework}
\label{subsec:framework}

The framework is currently written from scratch in C++ with our intention to move to a framework such as Robot Operating System (ROS) when we upgrade the hardware next year. Currently there are two main loops:

\begin{itemize}
  \item \emph{Vision} -- Each cycle of the vision loop is synced with the input of the camera at 30 fps (our main source of new information about the environment). This loop processes the camera input and plans future tasks for the robot as discussed in Section~\ref{subsec:strategy}.
  \item \emph{Hardware} -- This thread operates at 100+ ups, servicing each motor and collecting non-camera inputs. This loop is updated by the \emph{vision loop} which requests an action relative to the current state/location observed.
\end{itemize}

\subsection{Vision}
\label{subsec:vision}

\begin{figure}
  \centering
  \includegraphics[width=\linewidth]{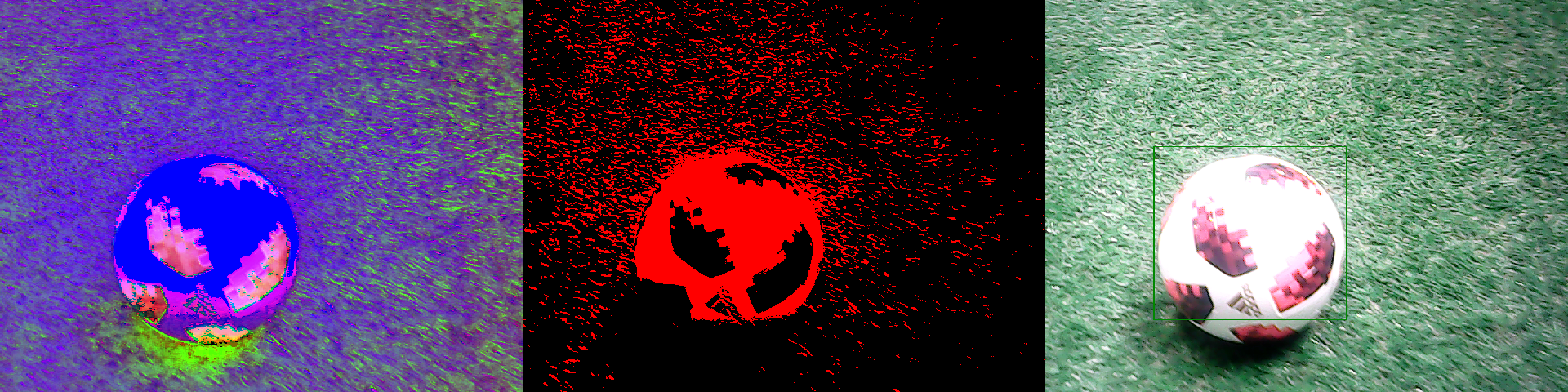}
  \caption{Left: HSV, Middle: labeling, Right: classification.}
  \label{fig:vision-example}
\end{figure}

\begin{figure}[H]
  \centering
  \begin{tikzpicture}
    [node distance=.3cm, start chain=going below]
    \node[punktchain, join] (b1) {Raw Image};
    \node[punktchain, join] (b2) {HSV Colour Space};
    \node[punktchain, join] (b3) {Horizon + Field Edge Filter};
    \node[punktchain, join] (b4) {Integral Image Candidate Selection};
    \node[punktchain, join] (b5) {Classification Filter};
  \end{tikzpicture}
  \caption{Vision pipeline overview.}
  \label{fig:vision-pipeline}
\end{figure}
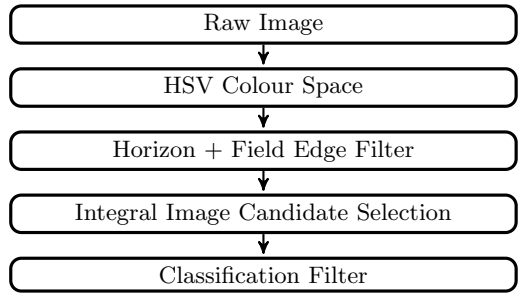

The robot uses a USB web camera which gives it an 640x480 pixel image in YUYV422 at 30 fps (see Figure~\ref{fig:vision-example}). Figure~\ref{fig:vision-pipeline} shows the order in which we process an image, where several basic algorithms are used to pre-process the image, with the purpose of reducing the search space for performing a classification filter, where a proposed ball candidate is rated on expected features such as: roundness, colour profile, distance, size and distance from previously identified location.
Figure~\ref{fig:vision-example} shows the results of the ball classification branch. Section~\ref{subsec:vision-research} specifies how we plan to extend this system in the near future.

We are currently using three classification filters: number of pixels, ratio of the pixels in the candidate area and the size of the candidate area. As the vision gets further developed, additional filters will be specified. However, the current three can successfully identify a white ball on the field under partial natural lighting conditions.

\subsection{Walking}
\label{subsec:walking}

Our walking algorithm is adopted from the Rhoban team (2012 - present) who described their method \textit{IKWalk} in depth in their 2015 Open Source Contribution paper \cite{rouxel2015rhoban}. Their walk engine is designed for small humanoid robots with 12 DoF in the legs. It generates an oscillatory pattern that defines parameterised trajectories, then computes the target motor positions for inverse kinematics. Unlike the Rhoban team, our walking engine does not receive pressure feedback from pressure sensors in the feet, but instead we compute offsets on walk engine parameters based on the predicted centre of mass from the IMU.

\subsection{Strategy}
\label{subsec:strategy}

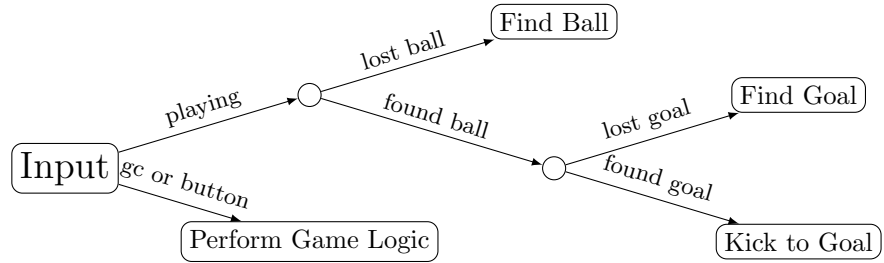
\begin{figure}[H]
  \centering
  \begin{tikzpicture}
    [grow = right, sibling distance = 6em, level distance = 10em, edge from parent/.style = {draw, -latex}, every node/.style = {font=\footnotesize}, sloped]
    \node [root] {Input}
      child {
        node [env] {Perform Game Logic}
        edge from parent node [above] {gc or button}
      }
      child {
        node [dummy] {}
        child {
          node [dummy] {}
          child {
            node [env] {Kick to Goal}
            edge from parent node [above] {found goal}
          }
          child {
            node [env] {Find Goal}
            edge from parent node [above] {lost goal}
          }
          edge from parent node [above] {found ball}
        }
        child {
          node [env] {Find Ball}
          edge from parent node [above] {lost ball}
        }
        edge from parent node [above] {playing}
      };
  \end{tikzpicture}
  \caption{Simplified robot gameplay behaviour.}
  \label{fig:behaviour}
\end{figure}

In Figure~\ref{fig:behaviour} we express a simplified behaviour tree for game-play which allows us to reason about the actions taken by the robot.

Team communication happens via the mixed team communication protocol \cite{FUmanoids2018}, and they are alsoe able to the game controller \cite{Bhuman2018}.

In order to detect a fall, gyroscope and accelerometer values are read where it is determined whether correctional actions are still viable. If the robot is unable to correct, it braces for impact by relaxing motors which reduces wear on the internal gears. Next a sequence of actions are run to get the robot safely into a known stance, after which the get up sequence is run.

\section{Research Interests}
\label{sec:research}


\subsection{Low-Cost Humanoid Research Platform}
\label{subsec:low-cost}

The purpose of creating a new humanoid robot platform was to do so with a lower entry budget, where each unit costs approximately \$2000 USD with the use of a moderately good 3D printer (such as the Makerbot series). Our intention is to open source the humanoid robot platform after the competition to encourage more teams to easily enter the competition. We believe this may also allow researchers to build multiple humanoid robots at a lower cost, with full knowledge and control over the system. One such use is discussed in Section~\ref{subsec:hri}.

\subsection{Localisation}
\label{subsec:localisation}

Our intention is to implement a simplistic localisation based on a heading derived from the gyroscope, as well as the location of the robot to be derived from edge of the field and other players reporting their position relative to the field and ball. This data will be fused and passed through an extended Kalman filter, where we expect to derive the approximate location and rotation of the player \cite{Kalman1960}. This will be tracked over time, where confidence is increased both upon continued matches compared to estimation and is calibrated on high confidence events, such as kick-off.

We are considering breaking the symmetry of the field through cooperative localisation. It would work through the use of the team communication protocol and would be consensus based on agent confidence values. Each agent would communicate their own confidence on where they are at in the field, and update their position based on the information obtained from the others.

\subsection{Human Robot Interaction (HRI)}
\label{subsec:hri}

The Human Interface Technology lab (HITlab NZ) has a number of people working on HRI research; specifically, social HRI (see for example \cite{zlotowski2013more,juarez2011using,brandstetter2017robots}). Even when well aware that robots are machines without consciousness, humans still automatically interact with them as if they were social agents \cite{krach2008can,rosenthal2013neural}. This includes them expecting robots to adhere to social norms \cite{eyssel2012activating,mutlu2008robots}, like turn-taking in a conversation and stepping to the side on time when their pathways are about to collide; and then planning their own behaviour based off of these expected robot behaviours. In terms of the RoboCup, this means that if robots and humans are ever going to play a football game together, robots will not just have to be able to play football but to play football in the same ways as humans.

In the shorter term, developing a cheap robot platform allows the HITlab NZ to customize the robot (Section~\ref{subsec:low-cost}) to the experiment at hand up to a certain extent. Moreover, with one of the PhD projects focusing on robot abuse, developing a low cost platform allows for designing experiments where participants can potentially damage the robots they're interacting with, since repairing the robot and replacing any damaged parts will be relatively cheap and replacement parts can be purchased or printed in advance.

\subsection{Bipedal Motion Engine}
\label{subsec:motion}

Our current walk engine consists of a custom scripted gaited bipedal walk, where various predetermined parameters (such as stride length, timing and joint offset) are used to actively balance the robot with the use of the gyro to determine how much correction is needed. Better well known and tested methods also exist for bipedal walking such as zero moment point (ZMP), where the center of pressure (CoP) is to remain within the support polygon during movement \cite{VukobratovicBorovac2004}.

Unfortunately, like many other teams, this still leaves procedures such as ``get-up from fallen" to be scripted, sometimes resulting in the robot unable to recover from some scenarios. In order to move towards our goal in 2050, our robots must be able to do more than slow walking to compete with human players and will have to do so in much more challenging settings.

We intend to experiment with specifying an end-target pose as an input and to a \emph{motion engine}, which computes a pose for the next time step. Instead of specifying a sequence of actions for the robot to stand-up, we instead internally compute actions that help achieve our goal. Initial validation will be done using the MuJoCo physics engine \cite{TodorovErezTassa2004}.

\subsection{Vision}
\label{subsec:vision-research}
We propose that to detect a ball and goal, we adopt and improve the methods proposed in \cite{SusantoRudiawanAnaliaSutopoSoebakti2017} and \cite{KimParkLeeOh2005}.

\begin{itemize}
  \item \emph{Ball Detection:} To detect the ball, we plan to use an edge detection algorithm in combination with blob detection and filling algorithm to get several candidate blobs. The detected blob candidates are then passed to the ball recognition system, which is a pre-trained convolutional neural network (CNN). We also plan to use an extended Kalman filter to track the ball and this information is then integrated into the ball recognition system in the subsequent image sequences. We are actively in the process of improving both accuracy and computational efficiency of the ball recognition system using Basic Target Searching Algorithm, as proposed in \cite{KimParkLeeOh2005}.
  \item \emph{Line Detection:} To detect field and goal lines, we intend to implement the method proposed by \cite{FaraziAllgeuerBehnke2016}. More specifically, an edge detector followed by probabilistic Hough line detection is to be applied. Smaller segments are to be  filtered to avoid false positives. Finally, the remaining similar line segments are merged by using a straight-line algorithm to have fewer larger and accurate lines. The shorter line segments are used for detecting the field circle, while the remaining lines are passed to the localization method.
  \item \emph{Goal Detection:} We are trialling two methods to detect the goal post. The first method is to have a specific pattern on our goalkeeper and detect this, as well as goal lines, to differentiate our goal posts from the opponent goal posts. The second method involves a complete object recognition model using CNN to detect the goal in localised regions by means of a line detection algorithm.
\end{itemize}

\section{Acknowledgements}
\label{sec:acknowledgements}

We would like to acknowledge the open source projects: OpenSCAD and GitLab. We would also like to thank the University of Canterbury, specifically the College of Engineering, the School of Product Design, and the HITlab NZ, for providing financial and material resources. Finally, we'd like to thank the following persons: Associate Professor Christoph Bartneck, Associate Professor Richard Green, Dr Simon Hoermann and Dr Tim Huber.

\bibliographystyle{splncs04}
\bibliography{bibliography}
\end{document}